\documentclass{article}

\usepackage[ruled,vlined]{algorithm2e}%
\usepackage{arxiv}
\usepackage[square,numbers]{natbib}
\usepackage[utf8]{inputenc} 
\usepackage[T1]{fontenc}    
\usepackage{hyperref}       
\usepackage{url}            
\usepackage{booktabs}       
\usepackage{amsfonts}       
\usepackage{nicefrac}       
\usepackage{microtype}      
\usepackage{lipsum}		
\usepackage{graphicx}
\usepackage{doi}
\usepackage{tabularx}
\usepackage{graphicx}
\usepackage{subcaption}
\usepackage{amsmath}
\usepackage{mdframed}
\usepackage{longtable}
\newcolumntype{P}[1]{>{\raggedright\arraybackslash}p{#1}}
\usepackage{comment}
\usepackage{csquotes}
\usepackage[table]{xcolor}
\usepackage{multirow}
\usepackage{array}
\usepackage{listings}
\usepackage{rotating}   
\usepackage{colortbl}%
\usepackage{threeparttable}   

\usepackage{soul}
\usepackage{xcolor}

\usepackage{fancyvrb}

\definecolor{HeavyGreen}{HTML}{008000}  
\definecolor{Shamrock}{HTML}{03AC13}    
\definecolor{MidOrange}{HTML}{FFB000}   
\definecolor{BadRed}{HTML}{E02020}      
\definecolor{ShockBg}{HTML}{F8D7DA}   

\newcommand{\good}[1]{\textcolor{Shamrock}{#1}}   
\newcommand{\ok}[1]{\textcolor{HeavyGreen}{#1}}   
\newcommand{\meh}[1]{\textcolor{MidOrange}{#1}}   
\newcommand{\bad}[1]{\textcolor{BadRed}{#1}}      

\renewcommand{\good}[1]{\cellcolor{green!25}{#1}}   
\renewcommand{\ok}[1]  {\cellcolor{yellow!30}{#1}}  
\renewcommand{\meh}[1] {\cellcolor{orange!30}{#1}}  
\renewcommand{\bad}[1] {\cellcolor{red!25}{#1}}     

\title{Generalisation Bounds of Zero-Shot Economic Forecasting using Time Series Foundation Models}

\author{ 
  \href{https://orcid.org/0009-0003-7339-9294}{\includegraphics[scale=0.06]{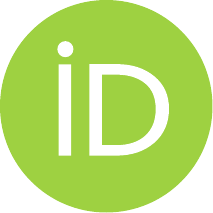}\hspace{1mm}Jittarin Jetwiriyanon}\thanks{Corresponding author: jittarin.jetwiriyanon.1@uni.massey.ac.nz} \\ 
  School of Mathematical and Computational Sciences\\%
  Massey University\\%
  Albany, New Zealand
  \and
  \href{https://orcid.org/0000-0001-9416-1435}{\includegraphics[scale=0.06]{orcid.pdf}\hspace{1mm}Teo Susnjak}\thanks{Contributing author: T.Susnjak@massey.ac.nz}\\%
  School of Mathematical and Computational Sciences\\%
  Massey University\\%
  Albany, New Zealand
  \and
  \href{https://orcid.org/0000-0003-0701-0204}{\includegraphics[scale=0.06]{orcid.pdf}\hspace{1mm}Surangika Ranathunga}\thanks{Contributing author: S.Ranathunga@massey.ac.nz}\\%
  School of Mathematical and Computational Sciences\\%
  Massey University\\%
  Albany, New Zealand
}

\renewcommand{\shorttitle}{Generalisation Bounds of Zero-Shot}

\begin{document}
\maketitle

\begin{abstract}
This study investigates the transfer learning capabilities of Time Series Foundation Models (TSFMs) under the zero-shot setup, to forecast macroeconomic indicators. New TSFMs are continually emerging, offering significant potential to provide ready-trained and accurate forecasting models that generalise across a wide spectrum of domains. However, the transferability of their learning to many domains, especially economics, is not well understood. To that end, we study TSFM's performance profile for forecasting economic, bypassing the need for training bespoke econometric models using extensive training datasets. Our experiments were conducted on a univariate case study dataset, in which we rigorously back-tested three state-of-the-art TSFMs (Chronos, TimeGPT and Moirai) under data-scarce conditions and structural breaks. Our results demonstrate that appropriately engineered TSFMs can internalise rich economic dynamics, accommodate regime shifts, and deliver well-behaved uncertainty estimates out of the box, while matching and exceeding state-of-the-art multivariate models currently used on this domain. Our findings suggest that, without any fine-tuning and additional multivariate inputs, TSFMs can match or outperform classical models under both stable and volatile economic conditions. However, like all models, they are vulnerable to performance degradation during periods of rapid shocks, though they recover the forecasting accuracy faster than classical models. The findings offer guidance to practitioners on when zero-shot deployments are viable for macroeconomic monitoring and strategic planning.
\end{abstract}


\section{Introduction}
Macroeconomic indicators, such as gross domestic product (GDP), consumer price inflation, and the unemployment rate, serve as gauges for the economy's direction. These indicators distil vast amounts of data into manageable signals of aggregate demand, supply-side capacity, and financial conditions. Timely and reliable forecasts of gross domestic product (GDP) are essential inputs for policy, finance, and corporate planning worldwide. Governments embed medium-term growth scenarios in budget frameworks, debt-sustainability analyses, and multi-year spending envelopes \cite{IMF_WEO_2024}. Prudential regulators feed GDP paths into system-wide stress tests to gauge the resilience of banks and insurers under adverse macroeconomic conditions \cite{BorioDrehmann2014}. Investors rebalance portfolios when growth expectations shift, treating GDP as a succinct proxy for business-cycle momentum \cite{Bloom2014}. Multinational firms align production schedules, inventories, and staffing levels with headline and sectoral GDP projections \cite{OECD_EO_2023}, while credit-rating agencies and development institutions incorporate forward-looking GDP assumptions into sovereign-risk assessments and concessional-finance formulas \cite{SP_Sovereign_Methodology_2017}.

However, producing dependable forecasts remains challenging: structural breaks, measurement errors, and sudden shocks can quickly erode model performance, thereby motivating continuous innovation, from classical econometric combinations to modern machine-learning approaches \cite{ClarkWest2007}. Forecasting GDP accurately across countries presents significant challenges. Firstly, early data releases are frequently revised, sometimes to the extent of reversing the sign of reported growth. This means models must contend with evolving truths rather than a fixed target \cite{CroushoreStark2001}. Secondly, structural changes, such as those triggered by commodity-price super-cycles, financial crises, natural disasters, and global pandemics, can abruptly disrupt historical relationships and invalidate previously stable parameters \cite{Perron1989,Hamilton2009Oil,CavalloNoy2011,JordaSinghTaylor2022Pandemics}. Lastly, sector-level GDP series display diverse seasonal patterns and react unevenly to external-demand shocks, hindering the transfer of information from one industry to another \cite{Marcellino2005Sector}.

For decades, classical time series models, such as autoregressive integrated moving-average (ARIMA) frameworks \cite{BoxJenkins1976} and vector autoregressions (VARs) \cite{Sims1980}, have been used as forecasting tools for this domain due to their transparency, tractability, and ease of re-estimation. However, their core assumptions of linearity and stable parameters rarely hold in practice, and sudden shifts in policy regimes and technology shocks can all invalidate coefficients calibrated on historical data, causing forecast accuracy to deteriorate rapidly as the horizon lengthens \cite{FildesStekler2002}. Empirical surveys have shown that beyond a few quarters ahead, even well-specified and rich multivariate systems seldom outperform simple persistence or random-walk benchmarks \cite{GiannonePredictability}. This challenge is further compounded by the constant moving of the goalposts in the form of data revisions, where preliminary GDP releases and other key indicators are often substantially updated, meaning models trained on early vintages chase a moving target and deliver the least reliability precisely when decision-makers most need clarity \cite{JacobsVanNorden2011}.
Recent studies have explored more sophisticated modelling approaches such as mixed-frequency factor and Bayesian models, which integrate hundreds of monthly indicators to sharpen nowcasts \cite{Carriero2020MixedFreq}, as well as tree-based ensembles and hybrid neural networks capable of discovering non-linear interactions \cite{Maccarrone2021MLGDP}.
However, the more recent modelling approaches are accompanied by high overheads in terms of expertise and computation for modest returns in improved accuracies, while being dependent on provision and access to real-time multivariate inputs \cite{long2023scalableprobabilisticforecastingretail}.

Recent advancements in AI have introduced a new class of forecasting tools, namely Time Series Foundation Models (TSFMs), with the potential to mitigate longstanding challenges in macroeconomic forecasting \cite{Goel2025VaR}. TSFMs attempt to leverage transfer learning, which reuses representations learned on a source dataset in order to re-purpose it for making forecasts on a target task. TSFMs are pre-trained on millions of heterogeneous time series with knowledge captured in parameters that in theory can then be transferred to improve performance on a different target either with or without model fine-tuning \cite{GermanMorales2025}. Under the zero-shot  regime, no further fine-tuning is undertaken, and thus the transferability of TSFMs to accurately forecast future values can then be explored in the purest form, thus significantly reducing the modelling resource overheads \cite{Bommasani2022}.
These large pre-trained TSFMs, Nixtla's TimeGPT, Amazon's Chronos, and Salesforce's Moirai treat numeric sequences as language tokens and leverage transformer backbones trained on extensive datasets. TimeGPT offers a ``plug-and-play'' API capability providing zero-shot forecasts without local fine-tuning, while Moirai extends this paradigm to multivariate forecasting with exogenous covariates. Models like Chronos have been used in literature on domains such as electricity, traffic, and retail data, where they have reported better performances than tuned statistical baselines with minimal feature engineering \cite{TimeGPT2024, Moirai2024, Chronos2024}.
Given the largely unexplored capabilities of these models in a macroeconomic context, our study seeks to explore to what degree the current cutting-edge TSFMs can generalise across data vintages, sectors, and structural breaks in economic datasets, under the ``out-of-the-box'', zero-shot settings. For our study, we used macroeconomic data from New Zealand as a case study, which presents an unusually demanding testbed, given the vulnerability of its small economy to external factors, as well as its exposure to commodity-price cycles, natural disasters, and external-demand shocks \cite{McKenzie2024}, together with the frequent revisions and thus the uncertainty of macroeconomic indicators estimates \cite{WestpacGDPRevisions2024}.

\subsection*{Contribution and Novelty}
While acknowledging that optimal macroeconomic forecasting ideally requires the integration of rich, real-time features capturing key economic indicators as model inputs, this study adopts a distinct approach. We focus specifically on evaluating the pure zero-shot forecasting capabilities of TSFMs demonstrate true zero-shot transfer learning without incorporating such external features or domain-specific covariates. This deliberate choice is made to establish fundamental performance limits and explore the inherent forecasting potential of these models based solely on their pre-training, thereby establishing baselines of their transfer abilities, given the current paucity of research applying TSFMs, particularly in a zero-shot manner, to the economic domain. The contributions and novelty of this work can be summarised as:

\begin{itemize}
    \item \textbf{Empirical benchmark}: we provide the zero-shot evaluation of leading TSFMs (Chronos, Moirai, TimeGPT) against classical econometric baselines from the Reserve Bank of New Zealand (RBNZ) forecasts, covering New Zealand’s national GDP and sectoral industries.
    \item \textbf{Performance measurement}: we demonstrate that TSFMs outperform other classical methods across various horizons, including RBNZ’s benchmark models, and thus we establish their utility under certain conditions.
    \item \textbf{Operational guidance}: we offer actionable insights for policy analysts by mapping the boundary conditions under which zero-shot TSFMs serve as low-maintenance forecasting tools for practitioners or economists. We also identify scenarios where lightweight classical models remain preferable.
\end{itemize}

\section{Related works}\label{sec2}
\subsection{Forecasting Difficulty for Macroeconomic Indicators}
Forecast accuracy for macroeconomic aggregates is fundamentally constrained by low signal-to-noise ratios. A long tradition of forecast evaluation studies shows that, once the horizon stretches beyond the \textit{now-cast} and the subsequent quarter, point predictions of real GDP growth seldom beat a naive random walk, let alone a purely random-direction guess, when measured against out-of-sample performance \cite{GiannonePredictability}. Persistence forecasts, which simply carry forward the latest observed growth rate, provide a standard benchmark over which, even multivariate econometric systems rarely improve upon this baseline after the first step in the forecast horizon \cite{Hartigan2024RDP}. Even median private-sector projections at a four-quarter horizon exhibit RMSEs statistically indistinguishable from persistence \cite{EdgeRudd2016}.

This bound tightens whenever rare shocks such as financial crises, pandemics, natural disasters, and geopolitical conflicts create unexpected changes that historical data cannot anticipate. Empirical work documents steep declines in forecast performance during such events \cite{Castle2011ForecastBreaks}; the COVID-19 pandemic, for example, overwhelmed both sophisticated econometric models and advanced machine-learning systems because existing training sets contained no historical analogue \cite{Lewis2022WEI}. This shortcoming has motivated a shift toward non-linear and high-dimensional techniques. Machine-learning ensembles (random forests, gradient boosting), support-vector regression, and penalised regressions demonstrate gains by exploiting rich predictor sets \cite{RossiGuhathakurta2023Review}. Deep networks extend those gains: LSTMs beat tuned ARIMA baselines in volatile GDP series \cite{Oancea2024LSTM}, while residual architectures such as N-BEATS win open forecasting competitions when data are plentiful or creatively augmented \cite{Oreshkin2020NBEATS}.

To complement algorithmic advances, researchers now emphasise the timing of data arrival. Mixed-frequency and real-time approaches integrate high-frequency indicators, electronic-card transactions, and daily mobility into quarterly GDP nowcasts \cite{susnjak2018nowcasting}, providing policymakers near-instant feedback during shocks \cite{GiannoneRealTime,Herculano2022FCI}. Model builders also adapt specifications to the country context. In open economies like New Zealand, global commodity prices, foreign demand, and idiosyncratic domestic cycles jointly shape growth dynamics; assessing how well models internalise these influences remains an active line of inquiry \cite{Galt2000}.

Today, the cutting edge is TSFMs pre-trained on millions of heterogeneous sequences which promise a further step change. Offering zero-shot and few-shot forecasts with native probabilistic outputs, TSFMs circumvent manual indicator selection and merge deep-learning pattern discovery with classical uncertainty quantification. However, whether this architecture can overcome the persistence benchmark in shock-prone, data-sparse settings such as New Zealand is still an open question, and therefore, the specific gap that the present study intends to address.

\subsection{Modern and Emerging Forecasting Approaches}
Economic linear models were the focus of early GDP-forecasting research. Single-equation and small-VAR frameworks inherently assume linear relationships, potentially missing signals within extensive indicator sets. Although Bayesian shrinkage (BVAR) aids in preventing overfitting in larger VARs \cite{Wu2015VARHandbook, Litterman1986}, the linear assumption can be a significant limitation.

To harvest that broader information set, the literature turned to large-information factor techniques. Dynamic factor models (DFMs) compress hundreds of macro-financial series into a handful of latent factors that feed simple forecasting equations, delivering substantial accuracy gains \cite{StockWatson2002}. Central banks now view DFMs or their extensions as baseline nowcast engines: FAVARs embed factors inside VAR structures for structural analysis \cite{Bernanke2005FAVAR}, MIDAS regressions link monthly factors to quarterly GDP for real-time monitoring \cite{Ghysels2007MIDAS}, and document a principal-component DFM as the Reserve Bank of New Zealand’s benchmark tool \cite{Pick2024}.

Building on these statistical platforms, institution-specific suites provide operational nowcasts. The Federal Reserve Bank of New York’s medium-scale DSGE integrates theory-consistent shocks with factor information for policy analysis \cite{DelNegro2014DSGE}. In contrast, the Atlanta Fed’s GDPNow decomposes each GDP sub-aggregate via bridge equations and updates almost daily, offering a transparent, additive view of U.S. growth \cite{Higgins2014GDPNow}.
More recently, attention has shifted to machine learning and hybrid methods that relax linearity and exploit high-dimensional features. Ensemble trees (random forests, gradient boosting) already outperform factor and penalised-regression baselines on Dutch GDP nowcasts \cite{Pick2024}. Deep neural networks, especially Long Short-Term Memory (LSTM) models, capture nonlinear temporal dependencies and have beaten ARIMA benchmarks in volatile settings \cite{OanceaSimionescu2024ECECSR}. Hybrid ensembles push further by blending economic structure with ML flexibility: weighting forecasts from a time-varying-coefficient DFM and a recurrent neural network by inverse MSE reduces U.S. GDP errors beyond either model alone. At the same time, broader model-averaging strategies remain popular for error reduction \cite{Longo2021EnsembleGDP}.
A clear progression from linear autoregressions through factor-based systems and institutional nowcasting dashboards to data-hungry ML hybrids, each stage addressing limitations exposed by the last and setting the stage for evaluating emerging foundation-model approaches.

\subsection{Zero-Shot Transfer Learning for Macroeconomic Forecasting}
Zero-shot transfer learning applies a pre-trained model to a novel domain or task without requiring additional training or fine-tuning on new data. In the macroeconomic context, a model that has been pre-trained by vast heterogeneous datasets of economic indicators, frequencies, and business regimes to forecast fresh and new series that were never part of the training. Historically, this has been a significant challenge for time series analysis. However, recent research indicates that with appropriate architectures and training methodologies, zero-shot transfer learning is now achievable in this field \cite{Oreshkin_Carpov_Chapados_Bengio_2021}. This means a model pre-trained on a diverse collection of time series can directly forecast unseen time series without any further adjustments. This approach, also known as zero-shot forecasting, relies on the model's ability to capture universal temporal patterns that transfer effectively to new series. Suppose the characteristics of the new series are adequately represented by the diverse patterns learned during pre-training. In that case, the model can generate reasonable forecasts without specific training for each new series \cite{dooley2023forecast}.
Given a new domain that was never seen during training to perform inference without gradient updates \cite{zhou2023fitsallpowergeneraltime}.

\begin{equation}
\hat{\mathbf y}_{t+1:t+H}
= F_{\theta^\star}\bigl(\mathbf y^\dagger_{1:t},\,\mathbf x^\dagger_{1:t+H}\bigr)
\label{eq:zeroforecasting}
\end{equation}

Equation \ref{eq:zeroforecasting} describes a point of forecast for the next \(H\) time steps, using only the past \(t\) observations \(\mathbf y^\dagger_{1:t}\) and any covariates \(\mathbf x^\dagger_{1:t+H}\), without any parameter updates, provides the forecast of the distribution over the future \(H\) prediction \cite{auer2025zeroshottimeseriesforecasting}.

The feasibility of this approach has significantly increased with the emergence of TSFMs. These are large models trained on heterogeneous time series data from numerous domains. The benefits of zero-shot transfer learning are substantial, it eliminates the need for task-specific training and avoids the requirement for large target datasets. Furthermore, zero-shot forecasting specifically focuses on predicting future values for new time series by leveraging knowledge from a broad pre-trained model, treating each new time series as a zero-shot task. Recent studies have demonstrated surprisingly strong results for zero-shot forecasting using pre-trained models \cite{xiao2025timefoundfoundationmodeltime}.

\subsection{Zero-Shot TSFMs in Economic Forecasting}
That early evidence transformers can act as generic sequence learners came from the Frozen Pre-trained Transformer (FPT) experiment, where a language-model backbone was kept entirely frozen across diverse time series tasks, showing that self-attention can operate as a domain computation \cite{Zhou2023OneFitsAll}. Follow-up work systematises this line of research by mapping architectures, pre-training objectives, adaptation strategies, and data modalities \cite{Liang_2024}. While architectural variants such as encoder–decoder hybrids \cite{zhou2021informerefficienttransformerlong}, sparse-attention blocks \cite{beltagy2020longformer}, and decomposition-style residual paths \cite{wu2022autoformer} dominate now with networks of millions of parameters that are pre-trained on a vast heterogeneous collection of time series. These TSFMs are typically transformer-based pre-trained models on massive collections of time series data, enabling zero-shot capabilities in forecasting in the concept of zero-shot transfer learning, which allows models to generalise to unseen tasks by leveraging knowledge gained from previously seen data \cite{yoon2022zeroshottransferlearningheterogeneous}.

\begin{figure}[htbp]
  \centering
  \includegraphics[width=0.9\textwidth]{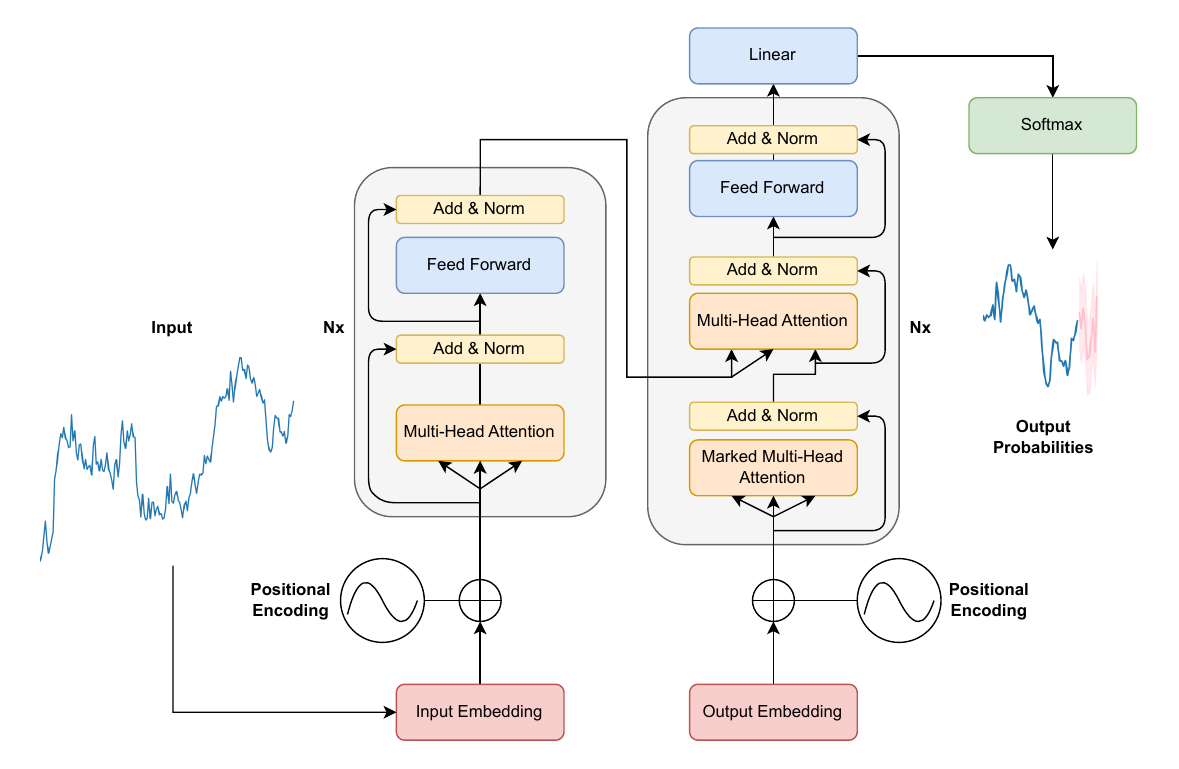}
  \caption{The transformer architecture. \cite{vaswani2017attention}}
  \label{fig:transformer_model}
\end{figure}

TSFMs are based on the transformer architecture, shown in Figure \ref{fig:transformer_model}. TSFMs aim for broad transferability with little or no task-specific fine-tuning. A recent survey \cite{ye2024surveytimeseriesfoundation} on TSFMs drew an important distinction between work that pre-trains transformers directly on raw time series data and work that adapts existing large-language-model (LLM) backbones. Recently, the emergence of several TSFMs illustrates how these architectural ideas are realised at scale:
\begin{itemize}
    \item Chronos repurposes the T5 language backbone for sequence-to-sequence forecasting, capturing fine-grained temporal dependencies \cite{Chronos2024}.
    \item Moirai pushes universality further by introducing multi-patch projections that sidestep fixed-frequency constraints and perform well on both sub-hourly energy usage and daily retail sales \cite{Moirai2024}.
    \item TimeGPT showed that a single globally trained network can forecast across hundreds of public datasets without per-task fine-tuning \cite{TimeGPT2024}.
\end{itemize}
Despite these successes, most comparative studies still rely on classic electricity-load (ETT), traffic, and retail (M5) datasets. Even recent transformer baselines are benchmarked on ETT and weather \cite{zhou2021informerefficienttransformerlong}. Consequently, it is not well understood from the literature whether the celebrated efficiency of TSFMs extends to the noisier, lower-sample-size realm of macroeconomic forecasting. Likewise, zero-shot experiments with numeric-token LLMs show promise on synthetic or industrial-sensor data \cite{JinTimeLLM, BrownLanguageModels}, but stop short of testing sturdiness under sudden shifts like the COVID-19 shock.

\subsection{Summary and Research Questions}
This study investigates zero-shot forecasting capabilities of TSFMs to achieve zero-shot transfer learning in a macroeconomic forecast, deliberately foregoing any domain-specific features. By directly applying out-of-the-box, pre-trained TSFMs, we establish fundamental performance baselines that reveal how effectively these models can leverage their learned representations to forecast economic time series and perform under major shocks, all without any fine-tuning, and utilise national GDP and industry sectors data as a representative case study for generalisable research questions.
\begin{itemize}
    \item \textbf{RQ1} How effective are state-of-the-art TSFMs for zero-shot univariate forecasting with zero-shot transfer learning of macroeconomic and industry-level time series? \label{RQ1}
    \item \textbf{RQ2} To what extent do zero-shot TSFMs forecast remain stable when confronted with periods of extreme volatility and significant economic disruption? \label{RQ2}
    \item \textbf{RQ3} Can zero-shot TSFMs match or surpass the published forecast accuracy of expert judgement models produced by central banks and international agencies?
    \label{RQ3}
\end{itemize}

\section{Methodology}\label{sec3}
\subsection{Dataset}
The dataset sourced from Stats NZ \cite{statsnz_gdp_dec2024_csv}, comprises a continuous quarterly time series of quarterly annual percentage change for four headline sectors: National GDP, Primary Industries, Goods-Producing Industries, and Services Industries, spanning 1999Q3 to 2024Q3. This 26-year horizon captures both routine seasonal rhythms and major unexpected changes for forecasting models. Interpreting the annual-growth metric is straightforward: National GDP aggregates all sectoral industries, while Primary Industries show pronounced seasonality, Goods-Producing Industries respond to global demand and investment cycles, and Services Industries mirror domestic consumption and conditions. Tracking these growth rates across the sample reveals regular business-cycle turning points and the sharp dislocations of the 2008 Global Financial Crisis and the 2020–2021 COVID-19 periods that pose particular challenges for traditional forecasting techniques.

\subsubsection{RBNZ Operational Dataset}
To eliminate look-ahead bias, alongside the public‐release series from Stats NZ, we incorporate the real-time GDP forecasts that the Reserve Bank of New Zealand (RBNZ) use in the quarterly percentage change for overall National GDP forecasts. Specifically, we contrast their \cite{Richardson2021Nowcast} Gradient Boosting (LSBoost) and Factor model projections.

\subsection{Baseline Models}
We benchmark TSFMs against four widely used baselines, the persistence model, ARIMA model, the least-squares boosting ensemble (LSBoost), and the factor model.

\subsubsection{Persistence model}
The persistence forecasting model is given by equation \ref{eq:persistence}.
\begin{equation}
\hat{y}*{t+h} = y_t \quad \text{for all } h \geq 1,
\label{eq:persistence}
\end{equation}

Here \(\hat{y}*{t+h}\) is the forecast at time \(t+h\), and \(y_t\) is the last observed value. Several studies have highlighted the practical utility and inherent limitations of the persistence model. As a prominent example, the impacts of selecting persistence forecasts as baseline references for evaluating forecasting systems. They found that persistence benchmarks substantially influence the assessment of more advanced models, particularly in contexts with strong seasonal patterns \cite{Barnes2020Persistence}.

\subsubsection{ARIMA model}
The Autoregressive–Integrated–Moving-Average (ARIMA) by Nixtla \cite{garza2022statsforecast} is an automatic process that employs a stepwise search procedure to identify optimal ARIMA and seasonal orders. This is achieved by combining unit-root tests for stationarity with the minimisation of information criteria. Models form a parsimonious yet expressive class for linear dynamics and remain a statistical baseline in forecasting
The seasonal form, denoted $\text{ARIMA}(p,d,q)\times(P,D,Q)$, combines non-seasonal and seasonal operators. The model acts as a dynamic statistical benchmark, adapting to each sector’s unique autocorrelation structure and seasonality patterns. It applies differencing to handle trend and drift components and fits candidate models using a maximum search space to maintain computational efficiency.

\subsubsection{LSBoost(Least-Squares Boosting)}
LSBoost is a gradient boosting method in algorithm \ref{alg:lsboost}, which minimises the squared-error (least-squares) loss, effectively performing a functional gradient-descent step in the space of functions in each boosting round. This machine learning ensemble method sequentially combines a finite set of weak learners to improve predictive performance for regression problems \cite{Friedman2001}. It works by fitting each new learner to the residual errors of the previous ones.

\begin{algorithm}[H]
\DontPrintSemicolon
\caption{LSBoost Algorithm \cite{app11052126}}
\label{alg:lsboost}
\KwIn{Training set $\{(x_i,y_i)\}_{i=1}^N$, number of iterations $M$}
\KwOut{Boosted predictor $F_M(x)$}

Define loss function:
\[
  L(y,F) \;=\; \tfrac{(y - F)^2}{2}
  \quad\text{and let }F_m(x)\text{ be the current regression model.}
\]

\textbf{Initialization:}\quad
$F_{0}(x) \;\leftarrow\;\bar y
  \quad(\bar y=\tfrac1N\sum_i y_i)$\;

\For{$m\leftarrow1$ \KwTo $M$}{
  Compute residuals:
  \[
    \tilde y_i \;\leftarrow\; y_i \;-\; F_{m-1}(x_i)
    \quad(i=1,\dots,N)
  \]
  Fit weak learner parameters $(\rho_m,\alpha_m)$ by
  \[
    (\rho_m,\alpha_m)
      \;=\;
      \arg\min_{\rho,\alpha}
      \sum_{i=1}^N
      \bigl[\tilde y_i \;-\; \rho\,h(x_i;\alpha)\bigr]^2
  \]
  Update model:
  \[
    F_m(x)
      \;\leftarrow\;
      F_{m-1}(x)\;+\;\rho_m\,h\bigl(x;\alpha_m\bigr)
  \]
}
\Return{$F_M(x)$}\;
\end{algorithm}

\subsubsection{Factor model}
Factor models are a class of statistical models designed to explain the co‐movement among a large panel of observed variables \(x_{it}\) and a smaller set of latent factors \(F_t\), represented by equation \ref{eq:static_factor}.
\begin{equation}
  x_{it} = \boldsymbol{\lambda}_i^\top F_t + e_{it}, 
  \qquad i = 1,\dots,N,\; t = 1,\dots,T,
  \label{eq:static_factor}
\end{equation}
where \(F_t\in\mathbb{R}^r\) with \(r\ll N\) captures the common dynamics, \(\boldsymbol{\lambda}_i\) are series‐specific, and \(e_{it}\) are errors that are allowed weak cross‐sectional and serial correlation.

In macroeconomic applications, it is often desirable to model the dynamic evolution of the latent factors. Dynamic factor models (DFMs) augment the static framework by equation \ref{eq:dynamic_factor}.
\begin{equation}
  F_t = \Phi_1 F_{t-1} + \dots + \Phi_p F_{t-p} + u_t,
  \quad u_t \sim \mathcal{N}(0,\Sigma_u),
  \label{eq:dynamic_factor}
\end{equation}
which embeds lead-lag relationships among economic indicators. This model strategy underlies modern nowcasting systems that blend hundreds of monthly and high-frequency indicators into real-time GDP estimates \cite{FORNI2004231}.

\subsection{Time Series Foundation Models}
We benchmark each foundation model against baseline models, providing comprehensive comparisons.

\subsubsection{TimeGPT‑1 model}
TimeGPT-1 is a large foundation model for time-series forecasting that adapts the Transformer architecture to temporal data. The model is trained at scale on a massive and diverse dataset of over 100 billion data points, which includes time series from various domains like finance, economics, healthcare, weather, IoT sensors, and web traffic. This extensive and heterogeneous dataset includes series with multiple seasonalities, different cycle lengths, various trends, and significant amounts of noise. Training on such a wide variety of time-series patterns allows TimeGPT-1 to generalise effectively. It can produce accurate zero-shot forecasts on entirely new time series without any specific re-training. For instance, the model can accurately forecast New Zealand's GDP and its industry sub-series without any fine-tuning or parameter adjustments.
Architecturally, the model uses an encoder-decoder Transformer with stacked self-attention blocks, residual connections, layer normalisation and positional encodings for the input window. A final projection layer maps the decoder's output to the desired forecast horizon, and the learned parameters are frozen to maintain its zero-shot forecasting capability \cite{TimeGPT2024}. This is demonstrated by direct comparison to baseline models.

\subsubsection{Chronos model}
Chronos is a pre-trained time series forecasting model that re-frames forecasting as a language modelling task. The model is trained on a large corpus of publicly available time series data from diverse sectors like energy, finance, and retail.
To enable this, Chronos first converts real-valued time series observations into a sequence of discrete tokens by using scaling and quantisation. It trains a transformer model (specifically, an off-the-shelf T5-style encoder-decoder) on these tokens using cross-entropy loss. This approach allows Chronos to learn generalised temporal representations that transfer across different datasets, resulting in enhanced forecasting accuracy. Forecasting proceeds by taking the final hidden states of the model for a given token sequence and, using a conditional distribution for the next token, autoregressively sampling new tokens. These tokens are then de-quantised to produce the zero-shot forecasting \cite{Chronos2024} to compare to baseline models.

\subsubsection{Moirai model}
Moirai, a masked-encoder universal Transformer, was developed by Salesforce AI Research for zero-shot forecasting. It operates across diverse time-series domains using the Large-scale Open Time-Series Archive (LOTSA), which includes datasets from nine domains: energy, transport, climate, cloud operations, web, sales, nature, economics/finance, and healthcare. To handle high-frequency series efficiently and reduce the quadratic cost of self-attention, the model uses larger patch sizes for embedding. It also introduces multiple patch-size input and output projection layers. The model employs an any-variate attention mechanism, which flattens multivariate time series into a single sequence. This mechanism utilises rotary position embeddings for time indices and learned binary attention biases for variate indices, optimising the negative log-likelihood by using the parameters of a mixture of distributions. Moirai is available in three sizes: Moirai-Small (6 layers, 14M parameters), Moirai-Base (12 layers, 91M parameters), and Moirai-Large (24 layers, 311M parameters).
Moirai provides competitive forecasts compared with fully pre-trained models, demonstrating the ability to transfer across domains and frequencies, using a single set of trained parameters to handle any domain and frequency in a zero-shot manner without fine-tuning or re-training. This approach allows Moirai to deliver flexible and accurate zero-shot forecasting on out-of-distribution datasets, handling diverse data from various sectors and exogenous covariates \cite{Moirai2024} for direct evaluation against baseline models.

\subsection{Model Evaluation}
In the forecasting literature, scale-dependent error measures remain fundamental for assessing forecast accuracy.

Mean Absolute Error (MAE) and Mean Squared Error/Root Mean Squared Error (MSE/RMSE) in equations \ref{eq:mse} and \ref{eq:rmse} are the two widely reported scale-dependent accuracy metrics. MAE specifically offers a direct interpretation as the average error in the original units of the series \cite{chatfield2000time}. Both compare a forecast \(\hat{y}_t\) against the observed value \(y_t\) over \(T\) time steps, but they weigh errors differently. MAE is the arithmetic mean of absolute deviations, squares the residuals before averaging, and then takes the square root. But their error weighting differs. MAE averages the absolute errors \(|y_t - \hat{y}_t|\), whereas RMSE averages the squared errors \((y_t - \hat{y}_t)^2\) before taking the square root.
\begin{equation}\label{eq:MAE}
    \text{MAE} \;=\; \frac{1}{T}\sum_{t=1}^{T} \lvert y_t - \hat{y}_t\rvert
\end{equation}
MAE makes it easy to interpret the forecast on average.

\begin{equation} \label{eq:mse}
\text{MSE}  = \frac{1}{T}\sum_{t=1}^{T}(y_t - \hat{y}_t)^2
\end{equation}

\begin{equation} \label{eq:rmse}
\text{RMSE} = \sqrt{\text{MSE}}
             = \sqrt{\frac{1}{T}\sum_{t=1}^{T}(y_t - \hat{y}_t)^2}.
\end{equation}

MSE/RMSE as defined in equations \ref{eq:mse} and \ref{eq:rmse} squares errors, giving disproportionate weight to large deviations and making the metric sensitive to outliers or significant irregularities. Both share a limitation: they are expressed in the original data units, preventing direct comparison across series with different scales. To address this, Symmetric Mean Absolute Percentage Error (SMAPE) in equation \ref{eq:SMAPE} normalises the absolute error by the average magnitude of the actual and forecast values \cite{Kim2016MAAPE},
\begin{equation}\label{eq:SMAPE}
    \text{SMAPE} \;=\; \frac{100}{T}\sum_{t=1}^{T} \frac{2\,\lvert y_t - \hat{y}_t\rvert}{\lvert y_t\rvert + \lvert \hat{y}_t\rvert}
\end{equation}
Expressed as a percentage, SMAPE offers direct comparison across series of different units or magnitudes and bounds the error between 0\% and 200\%. Nevertheless, SMAPE can become unstable if both \(y_t\) and \(\hat{y}_t\) approach zero simultaneously. Furthermore, arithmetic symmetry does not guarantee true statistical symmetry when distributions are highly skewed.
MASE in equation \ref{eq:MASE} provides an alternative approach for scale-free comparison. It also enables meaningful benchmarking against a naive baseline by dividing the MAE of the candidate model by the in-sample MAE of a simple seasonal naive forecast \cite{HYNDMAN2006679},

\begin{equation}\label{eq:MASE}
    \text{MASE} \;=\;
    \frac{\displaystyle\frac{1}{T}\sum_{t=1}^{T}\lvert y_t - \hat{y}_t\rvert}
     {\displaystyle\frac{1}{T-m}\sum_{t=m+1}^{T}\lvert y_t - y_{t-m}\rvert}
\end{equation}
where \(m\) is the seasonal period. This scaling provides a key advantage: MASE is a scale-free error metric. Because the denominator is the MAE of the naive seasonal forecast, the value serves as a direct benchmark. Thus, any MASE \(<\) 1 indicates performance superior to the naive baseline, and MASE \(>\) 1 denotes inferior accuracy. This scale-free property facilitates widespread adoption in comparative studies of forecasting algorithms.

The Diebold-Mariano (DM) in equations \ref{eq:DM1}, \ref{eq:DM2}, and \ref{eq:DM3}, introduced by \cite{DieboldMariano1995} established a general, loss-function-agnostic framework for testing whether two competing forecasts have the same expected predictive accuracy.
\begin{equation} \label{eq:DM1}
e_{i,t} = y_t - \hat{y}_{i,t}
\end{equation}
The forecast error from a model is \(i\), a loss function is \(\ell(\cdot)\), for example
\(\ell(e)=e^2\) (squared error) or \(\ell(e)=|e|\) (absolute error). 
Define the loss differential:
\begin{equation} \label{eq:DM2}
d_t = \ell\bigl(e_{1,t}\bigr) - \ell\bigl(e_{2,t}\bigr)
\end{equation}
and let
\(\bar d = \frac{1}{T}\sum_{t=1}^T d_t\).
For a \(h\)-step-ahead forecast, the DM statistic is

\begin{equation} \label{eq:DM3}
\mathrm{DM}
= 
\frac{\bar d}{\sqrt{\widehat{\mathrm{var}}(d_t)/T}},
\qquad
\widehat{\mathrm{var}}(d_t)
=
\gamma_0
+ 2\sum_{k=1}^{h-1}\Bigl(1-\tfrac{k}{h}\Bigr)\,\gamma_k,
 \end{equation}
 where \(\gamma_k\) represents the sample autocovariance of \(d_t\) at lag \(k\). Under suitable regularity conditions, \(\mathrm{DM}\) is asymptotically distributed as \(\mathcal N(0,1)\). Consequently, a two-sided \(z\)-test provides an asymptotic \(p\)-value. The use of the Newey–West (HAC) estimator in the denominator ensures the test's validity even when \(d_t\) follows a moving-average MA($h{-}1$) process, making it applicable interchangeably to MSE/RMSE or MAE comparisons.

\subsection{Zero-Shot Forecasts}
TSFMs such as TimeGPT, Chronos, and Moirai that employ zero-shot learning represent a significant advance in predictive analytics. Through large-scale pre-training on a wide variety of time series datasets, these models develop a generalised understanding of temporal dynamics, capturing regular cycles and anomalous events. Armed with this broad temporal intuition, they can be deployed directly on novel forecasting tasks without requiring extensive domain-specific fine-tuning.
At the core of zero-shot forecasting is the idea that, once a foundation model has internalised patterns spanning many industries and time scales, it can transfer that knowledge seamlessly to new contexts. This approach significantly reduces the time, computational resources, and specialised expertise normally needed for forecasting solutions, as it eliminates the need for extensive re-training and adaptation for each new task.

The practical benefits of zero-shot forecasting are most pronounced in settings where conditions change rapidly or data is scarce. For instance, in sectors such as National GDP, Primary Industries, Goods-Producing Industries, and Services Industries, unexpected events and economic shocks can frequently upend historical relationships and data patterns. In such dynamic environments, a zero-shot model can provide immediate, reasonably accurate projections without waiting for new data to accumulate or models to be re-trained. Furthermore, the same pre-trained model can often be utilised for both short-term operational decisions and longer-term strategic planning, all without needing to rebuild the model for each specific forecasting horizon.

\subsection{Experiment Pipeline}
This empirical experiment analyses the quarterly annual percentage change series for National GDP, Primary Industries, Goods‑Producing Industries, and Services Industries as Datasets. StatsNZ publishes the economic aggregates used in this analysis. The analysis employs a long window, spanning from 1999Q3 to 2024Q3, to expose the models to both secular growth phases and major shocks, including the Global Financial Crisis and the 2020–21 COVID collapse. Using a window expanding approach, accuracy is scored at every point with the evaluation metrics (MAE, RMSE, SMAPE, and MASE).

\begin{figure}[htbp]
  \centering
  \includegraphics[width=0.9\textwidth]{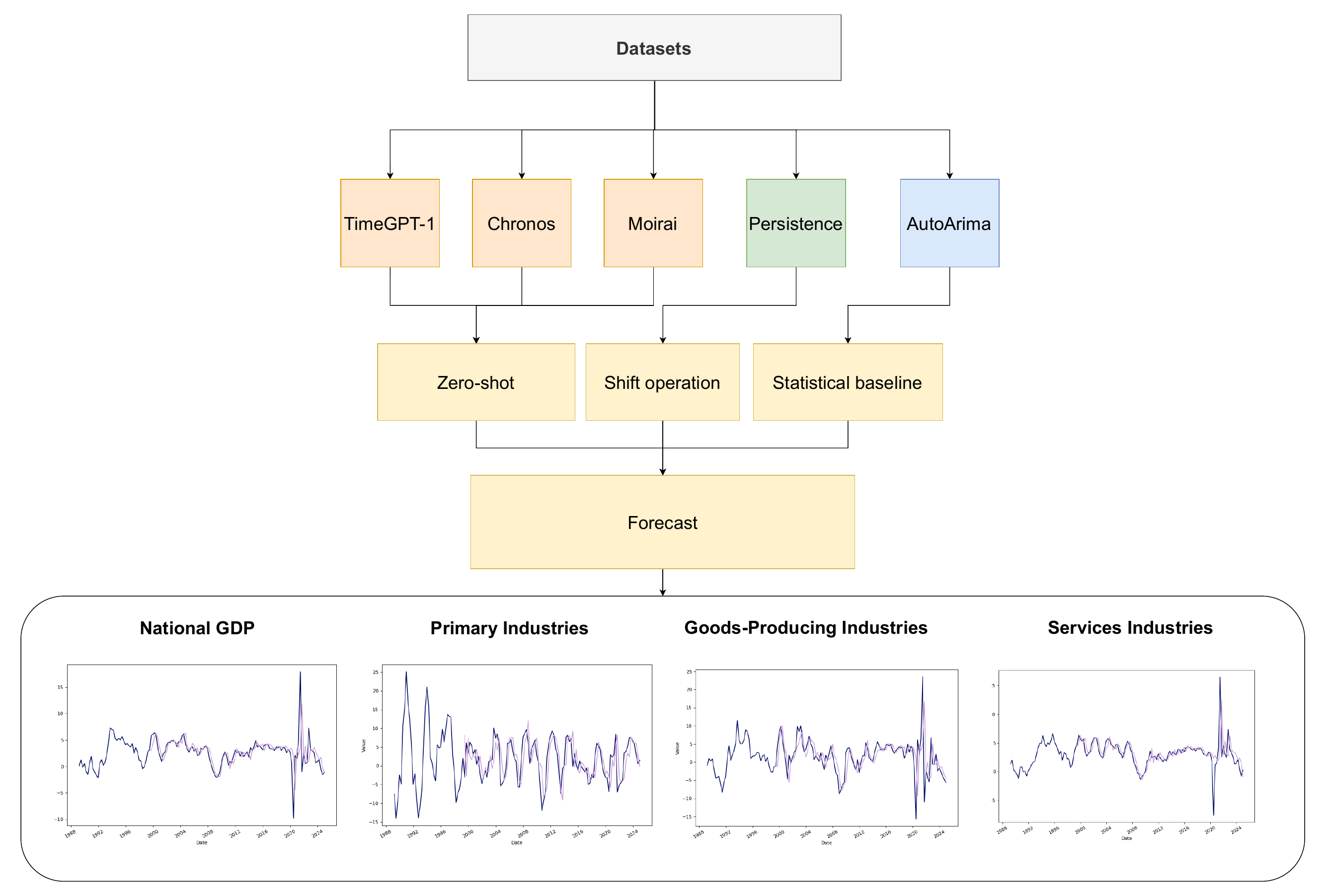} 
  \caption{The experiment workflow}
  \label{fig:experiment_Workflow}
\end{figure}

We design an experimental back-testing pipeline, shown in Figure \ref{fig:experiment_Workflow}. That compares shift operation persistence and statistical baseline ARIMA models against zero-shot forecasts from TSFMs: TimeGPT-1, Chronos-T5 (small, base, large), and Moirai (small, base, large). We generate one-quarter-ahead forecasts for each model over an expanding training window from 1999Q3 to 2024Q3. The forecasts are produced without any additional fine-tuning; the models were pre-trained before being evaluated on the same error metrics (MAE, RMSE, SMAPE, and MASE). To capture the heterogeneous dynamics of New Zealand's economy, we apply this experiment to four broad production sectors: National GDP, Primary Industries, Goods-Producing Industries, and Services Industries. The selection of these models for the experiments also reveals which classical and TSFMs are strongest with scarce data.

\section{Results}
\subsection{Analysis of Model Evaluation Results}
Table \ref{tab:forecast_stacked_vertical} benchmarks 9 forecasting models, including 7 TSFMs and 2 classical/statistical models, across four economic sectors: National GDP, Primary Industries, Goods-Producing Industries, and Services Industries. The analysis employs four accuracy metrics (MAE, RMSE, SMAPE, MASE) and four distinct time slices: a 26-year full sample, the calm Pre-COVID period, the COVID-19 shock years, and the Post-COVID rebound. Subsequently, Table \ref{tab:forecast_Diebold} details the Diebold-Mariano (DM) test results, comparing TSFMs with Persistence and ARIMA models.

\begin{sidewaystable}[p]
  \centering
  \resizebox{\textheight}{!}{%
    \begin{tabular}{l|cccc|cccc|cccc|cccc|c}
      \toprule
       & \multicolumn{4}{c|}{\textbf{National GDP}}
       & \multicolumn{4}{c|}{\textbf{Primary Ind.}}
       & \multicolumn{4}{c|}{\textbf{Goods-Prod. Ind.}}
       & \multicolumn{4}{c}{\textbf{Services Ind.}} & \textbf{Mean Rank} \\
      \cmidrule(lr){2-5}\cmidrule(lr){6-9}\cmidrule(lr){10-13}\cmidrule(lr){14-17}
      \textbf{Model} & MAE & RMSE & SMAPE & MASE
                     & MAE & RMSE & SMAPE & MASE
                     & MAE & RMSE & SMAPE & MASE
                     & MAE & RMSE & SMAPE & MASE & \textbf{RMSE} \\
      \midrule
      \multicolumn{17}{c}{\textbf{26-year Past-to-Present (1999Q3-2024Q3)}}\\
      \midrule
Persistence &
 \meh{1.44} & \bad{3.08} & \ok{0.49} & \meh{1.01} &
 \meh{3.02} & \meh{4.12} & \ok{0.88} & \meh{1.02} &
 \bad{3.06} & \bad{5.54} & \ok{0.90} & \bad{1.01} &
 \meh{1.18} & \bad{2.48} & \meh{0.38} & \meh{1.00} & 8.75 \\

Arima &
 \bad{1.49} & \meh{3.03} & \meh{0.50} & \bad{1.05} &
 \ok{2.39} & \ok{3.23} & \ok{0.76} & \ok{0.81} &
 \meh{2.97} & \ok{4.63} & \meh{0.93} & \meh{0.98} &
 \ok{1.15} & \meh{2.39} & \bad{0.39} & \ok{0.98} & 6.75 \\

TimeGPT-1 &
 \bad{1.49} & \ok{2.80} & \bad{0.52} & \bad{1.05} &
 \bad{3.32} & \bad{4.33} & \bad{0.99} & \bad{1.12} &
 \ok{2.93} & \meh{4.95} & \ok{0.91} & \ok{0.97} &
 \bad{1.20} & \ok{2.27} & \meh{0.38} & \bad{1.02} & 7.50 \\

Chronos-t5-small &
 \ok{1.22} & \ok{2.31} & \ok{0.49} & \ok{0.86} &
 \ok{2.70} & \ok{3.67} & \meh{0.90} & \ok{0.91} &
 \ok{2.71} & \ok{4.24} & \bad{0.94} & \ok{0.89} &
 \ok{1.01} & \ok{2.00} & \ok{0.36} & \ok{0.86} & 5.25 \\

Chronos-t5-base &
 \ok{1.23} & \ok{2.26} & \meh{0.50} & \ok{0.87} &
 \ok{2.76} & \ok{3.66} & \ok{0.89} & \ok{0.93} &
 \ok{2.70} & \ok{4.26} & \ok{0.92} & \ok{0.89} &
 \ok{0.98} & \ok{1.88} & \ok{0.36} & \ok{0.83} & 4.75 \\

Chronos-t5-large &
 \ok{1.38} & \ok{2.98} & \ok{0.49} & \ok{0.97} &
 \ok{2.78} & \ok{3.62} & \ok{0.88} & \ok{0.94} &
 \ok{2.79} & \meh{4.57} & \bad{0.94} & \ok{0.92} &
 \ok{1.02} & \ok{2.03} & \ok{0.37} & \ok{0.86} & 6.00 \\

Moirai-1.1-R-small &
 \ok{0.56} & \ok{1.49} & \ok{0.25} & \ok{0.40} &
 \ok{1.17} & \ok{1.57} & \ok{0.51} & \ok{0.40} &
 \ok{1.24} & \ok{3.05} & \ok{0.43} & \ok{0.41} &
 \ok{0.52} & \ok{1.31} & \ok{0.24} & \ok{0.44} & 2.75\\

Moirai-1.1-R-base &
 \good{0.48} & \good{1.04} & \good{0.20} & \good{0.34} &
 \ok{1.19} & \ok{1.60} & \ok{0.48} & \ok{0.40} &
 \good{0.91} & \good{1.57} & \ok{0.44} & \good{0.30} &
 \good{0.42} & \ok{1.18} & \good{0.17} & \good{0.36} & 1.75 \\

Moirai-1.1-R-large &
 \ok{0.52} & \ok{1.43} & \ok{0.22} & \ok{0.37} &
 \good{0.93} & \good{1.27} & \good{0.36} & \good{0.32} &
 \good{0.85} & \ok{1.86} & \good{0.36} & \good{0.28} &
 \ok{0.45} & \good{1.13} & \ok{0.20} & \ok{0.38} & 1.50 \\

\midrule
\multicolumn{17}{c}{\textbf{3-year Pre-COVID-19 (2017Q1-2019Q4)}}\\
\midrule

Persistence
 & \bad{0.41} & \meh{0.50} & \bad{0.13} & \meh{0.94}
 & \meh{2.29} & \meh{2.84} & \ok{0.90} & \meh{0.95}
 & \meh{1.19} & \meh{1.51} & \ok{0.39} & \ok{0.97}
 & \ok{0.33} & \meh{0.38} & \meh{0.10} & \ok{0.99} & 7.25 \\
Arima
 & \ok{0.39} & \ok{0.45} & \meh{0.12} & \ok{0.90}
 & \ok{1.80} & \ok{2.08} & \ok{0.71} & \ok{0.75}
 & \ok{1.13} & \ok{1.34} & \meh{0.41} & \ok{0.92}
 & \ok{0.27} & \ok{0.33} & \ok{0.08} & \ok{0.80} &  4.25 \\
TimeGPT-1
 & \bad{0.41} & \meh{0.51} & \meh{0.12} & \bad{0.95}
 & \bad{2.44} & \bad{2.99} & \bad{1.09} & \bad{1.02}
 & \ok{1.02} & \ok{1.39} & \ok{0.31} & \ok{0.83}
 & \bad{0.38} & \bad{0.44} & \bad{0.11} & \bad{1.13} & 7.75 \\
Chronos-t5-small
 & \meh{0.40} & \ok{0.47} & \bad{0.13} & \ok{0.92}
 & \ok{1.84} & \ok{2.37} & \meh{0.88} & \ok{0.76}
 & \ok{1.22} & \ok{1.52} & \meh{0.44} & \meh{0.99}
 & \ok{0.31} & \ok{0.34} & \ok{0.09} & \ok{0.91} & 6 \\
Chronos-t5-base
 & \meh{0.41} & \bad{0.55} & \bad{0.13} & \meh{0.94}
 & \meh{1.84} & \meh{2.43} & \meh{0.86} & \ok{0.76}
 & \meh{1.22} & \meh{1.53} & \bad{0.48} & \bad{1.00}
 & \ok{0.31} & \ok{0.35} & \ok{0.09} & \ok{0.92} & 7.5 \\
Chronos-t5-large
 & \ok{0.33} & \ok{0.40} & \ok{0.10} & \ok{0.75}
 & \meh{1.96} & \meh{2.54} & \meh{0.91} & \meh{0.82}
 & \ok{1.17} & \ok{1.45} & \ok{0.40} & \meh{0.95}
 & \meh{0.34} & \meh{0.38} & \meh{0.10} & \meh{1.00} & 6 \\
Moirai-1.1-R-small
 & \ok{0.26} & \ok{0.31} & \ok{0.09} & \ok{0.60}
 & \ok{0.94} & \ok{1.09} & \ok{0.54} & \ok{0.39}
 & \ok{0.32} & \ok{0.39} & \good{0.12} & \good{0.26}
 & \ok{0.23} & \ok{0.29} & \ok{0.07} & \ok{0.67} & 2.75 \\
Moirai-1.1-R-base
 & \ok{0.23} & \ok{0.29} & \ok{0.07} & \ok{0.52}
 & \good{0.70} & \good{0.83} & \good{0.28} & \good{0.29}
 & \ok{0.30} & \ok{0.40} & \good{0.12} & \good{0.25}
 & \good{0.11} & \good{0.14} & \good{0.03} & \good{0.33} & 1.75 \\
Moirai-1.1-R-large
 & \good{0.15} & \good{0.20} & \good{0.05} & \good{0.34}
 & \good{0.62} & \good{0.95} & \meh{0.36} & \good{0.26}
 & \good{0.18} & \good{0.25} & \meh{0.06} & \good{0.15}
 & \ok{0.14} & \ok{0.15} & \ok{0.04} & \ok{0.40} & 1.5 \\

\midrule
\multicolumn{17}{c}{\textbf{3-year During-COVID-19 (2020Q1-2022Q4)}}\\
\midrule

Persistence & \meh{6.42} & \bad{8.54} & \bad{1.32} & \meh{0.94} & \meh{3.92} & \bad{6.06} & \ok{0.75} & \meh{0.92} & \bad{11.13} & \bad{14.52} & \meh{1.43} & \bad{0.95} & \bad{5.11} & \bad{6.85} & \meh{0.94} & \bad{0.94} & 9 \\

Arima   & \bad{6.68} & \meh{8.35} & \ok{1.27} & \bad{0.98} & \ok{3.86} & \ok{4.83} & \meh{0.90} & \ok{0.90} & \meh{8.73} & \meh{10.98} & \ok{1.34} & \ok{0.75} & \meh{5.00} & \meh{6.56} & \bad{1.07} & \meh{0.92} & 6.5 \\

TimeGPT-1   & \ok{5.57} & \ok{7.30} & \ok{1.22} & \ok{0.82} & \bad{3.96} & \meh{5.69} & \bad{0.94} & \bad{0.93} & \ok{9.50} & \ok{12.22} & \bad{1.53} & \meh{0.81} & \ok{4.44} & \ok{5.86} & \ok{0.87} & \ok{0.82} & 7.25 \\

Chronos-t5-small & \ok{4.70} & \ok{6.21} & \meh{1.28} & \ok{0.69} & \ok{3.83} & \meh{5.46} & \ok{0.89} & \ok{0.90} & \ok{8.36} & \ok{10.28} & \bad{1.49} & \ok{0.72} & \ok{3.66} & \ok{5.34} & \ok{0.79} & \ok{0.67} & 5.25 \\

Chronos-t5-base  & \ok{4.53} & \ok{5.98} & \ok{1.25} & \ok{0.67} & \ok{3.76} & \ok{5.38} & \meh{0.91} & \ok{0.88} & \ok{8.60} & \ok{10.46} & \bad{1.55} & \ok{0.74} & \ok{3.42} & \ok{4.95} & \ok{0.81} & \ok{0.63} & 4.75 \\

Chronos-t5-large & \meh{6.00} & \ok{8.26} & \meh{1.28} & \ok{0.88} & \meh{3.92} & \ok{5.32} & \bad{0.94} & \meh{0.92} & \ok{9.24} & \ok{11.45} & \bad{1.60} & \ok{0.79} & \ok{3.75} & \ok{5.44} & \ok{0.79} & \ok{0.69} & 6.25 \\

Moirai-1.1-R-small & \ok{2.44} & \ok{4.17} & \ok{0.66} & \ok{0.36} & \ok{1.83} & \ok{2.21} & \ok{0.67} & \ok{0.43} & \meh{5.98} & \meh{8.59} & \ok{1.22} & \ok{0.51} & \ok{1.91} & \ok{3.60} & \ok{0.38} & \ok{0.35} & 3 \\

Moirai-1.1-R-base  & \good{2.06} & \good{2.83} & \good{0.60} & \good{0.30} & \good{1.40} & \good{1.78} & \good{0.48} & \good{0.33} & \good{2.73} & \good{3.78} & \good{0.64} & \good{0.23} & \good{1.65} & \good{3.26} & \good{0.35} & \good{0.30} & 1.5 \\

Moirai-1.1-R-large & \ok{2.54} & \ok{4.04} & \ok{0.77} & \ok{0.37} & \good{1.17} & \good{1.37} & \good{0.29} & \good{0.27} & \ok{3.41} & \ok{5.08} & \ok{0.76} & \ok{0.29} & \ok{1.81} & \good{3.15} & \ok{0.39} & \ok{0.33} & 1.5 \\

\midrule
\multicolumn{17}{c}{\textbf{2-year Post-COVID-19 (2023Q1-2024Q3)}}\\
\midrule

Persistence & \ok{0.85} & \ok{1.12} & \ok{0.71} & \ok{0.87} & \ok{1.53} & \ok{1.81} & \ok{0.43} & \ok{0.97} & \ok{1.71} & \ok{2.21} & \ok{0.86} & \ok{0.88} & \ok{0.89} & \ok{1} & \ok{0.68} & \ok{0.94} & 4.25 \\

Arima & \meh{1.66} & \meh{1.96} & \meh{1.18} & \meh{1.71} & \ok{1.53} & \ok{1.84} & \ok{0.38} & \ok{0.97} & \bad{4.71} & \bad{5.2} & \bad{1.73} & \bad{2.41} & \meh{1.24} & \meh{1.48} & \meh{0.75} & \meh{1.31} & 7.75 \\

TimeGPT-1 & \bad{2.46} & \bad{3.03} & \bad{1.21} & \bad{2.52} & \bad{2.31} & \bad{3.01} & \bad{0.65} & \bad{1.46} & \meh{3.85} & \meh{4.95} & \meh{1.4} & \meh{1.97} & \bad{2.16} & \bad{2.62} & \bad{0.98} & \bad{2.28} & 8.75 \\

Chronos-t5-small & \ok{0.94} & \ok{1.21} & \ok{0.82} & \ok{0.96} & \ok{1.17} & \ok{1.41} & \good{0.26} & \good{0.74} & \ok{2.13} & \ok{2.54} & \ok{1.28} & \ok{1.09} & \ok{0.87} & \ok{1.06} & \ok{0.66} & \ok{0.92} & 6 \\

Chronos-t5-base & \ok{0.92} & \ok{1.27} & \ok{0.78} & \ok{0.95} & \meh{1.59} & \meh{1.85} & \ok{0.33} & \bad{1} & \ok{2} & \ok{2.53} & \ok{1} & \ok{1.02} & \ok{0.89} & \ok{1.02} & \ok{0.68} & \ok{0.94} & 6.5 \\

Chronos-t5-large & \ok{0.87} & \ok{1.2} & \ok{0.76} & \ok{0.89} & \meh{1.9} & \meh{2.27} & \meh{0.64} & \meh{1.2} & \ok{2.03} & \ok{2.46} & \meh{1.12} & \meh{1.04} & \ok{0.89} & \ok{1} & \ok{0.69} & \ok{0.94} & 5.5 \\

Moirai-1.1-R-small & \ok{0.52} & \ok{0.6} & \good{0.42} & \ok{0.54} & \good{0.73} & \good{0.86} & \good{0.23} & \good{0.46} & \good{0.95} & \good{1.18} & \good{0.54} & \good{0.49} & \good{0.45} & \good{0.52} & \good{0.26} & \good{0.47} & 2 \\

Moirai-1.1-R-base & \ok{0.52} & \ok{0.62} & \ok{0.47} & \ok{0.53} & \ok{0.81} & \ok{1.18} & \meh{0.44} & \meh{0.51} & \good{0.57} & \good{0.69} & \ok{0.31} & \good{0.29} & \meh{0.65} & \meh{0.85} & \meh{0.38} & \meh{0.69} & 2.75 \\
Moirai-1.1-R-large & \good{0.43} & \good{0.55} & \ok{0.47} & \good{0.44} & \good{0.49} & \good{0.55} & \meh{0.36} & \meh{0.31} & \meh{0.58} & \meh{0.67} & \meh{0.46} & \ok{0.3} & \ok{0.49} & \ok{0.58} & \bad{0.52} & \ok{0.52} & 1.25 \\
    \bottomrule
    \end{tabular}}
  \caption{Forecasting results across four economic sectors in multiple horizons.}
  \label{tab:forecast_stacked_vertical}
\end{sidewaystable}

\begin{table}[htbp]
\centering
\scriptsize
\setlength{\tabcolsep}{3pt}
\begin{tabular}{l|cc|cc|cc|cc}
\toprule
\textbf{Model} & \multicolumn{2}{c|}{1999Q3-2024Q3} & \multicolumn{2}{c|}{Pre-COVID} &
\multicolumn{2}{c|}{COVID} & \multicolumn{2}{c}{Post-COVID}\\
\cmidrule(lr){2-3}\cmidrule(lr){4-5}\cmidrule(lr){6-7}\cmidrule(lr){8-9}
 & $p_P$ & $p_A$ & $p_P$ & $p_A$ & $p_P$ & $p_A$ & $p_P$ & $p_A$\\
\midrule
\multicolumn{9}{c}{\textbf{National GDP}}\\
TimeGPT-1            & 0.421 & 0.520 & 0.897 & 0.599 & 0.274 & 0.456 & 0.235 & 0.844 \\
Chronos\_Small       & 0.231 & 0.072 & 0.473 & 0.679 & 0.252 & 0.076 & 0.360 & 0.484 \\
Chronos\_Base        & 0.190 & 0.088 & 0.137 & 0.218 & 0.200 & 0.101 & 0.332 & 0.461 \\
Chronos\_Large       & 0.112 & 0.928 & 0.082 & 0.467 & 0.250 & 0.889 & 0.298 & 0.453 \\
\textbf{Moirai\_Small}& \textbf{0.048} & \textbf{0.017} & \textbf{0.033} & 0.154 & 0.087 & \textbf{0.044} & 0.126 & 0.187 \\
\textbf{Moirai\_Base} & \textbf{0.037} & \textbf{0.011} & \textbf{0.045} & 0.103 & 0.055 & \textbf{0.017} & 0.084 & 0.244 \\
\textbf{Moirai\_Large}& 0.061 & \textbf{0.022} & \textbf{0.016} & \textbf{0.012} & 0.104 & 0.052 & 0.097 & 0.225 \\
\addlinespace
\multicolumn{9}{c}{\textbf{Primary Industries}} \\
TimeGPT-1            & 0.275 & \textbf{0.001} & 0.686 & 0.088 & 0.119 & 0.614 & 0.102 & 0.156 \\
Chronos\_Small       & \textbf{0.005} & 0.054 & 0.327 & 0.320 & 0.277 & 0.559 & 0.296 & 0.866 \\
Chronos\_Base        & \textbf{0.012} & 0.057 & 0.455 & 0.274 & 0.389 & 0.630 & 0.472 & 0.656 \\
Chronos\_Large       & \textbf{0.020} & 0.067 & 0.594 & 0.209 & 0.298 & 0.687 & 0.979 & 0.414 \\
\textbf{Moirai\_Small}& \textbf{0.001} & \textbf{0.001} &
                         \textbf{0.030} & 0.076 & 0.144 & \textbf{0.035} & 0.238 & 0.300 \\
\textbf{Moirai\_Base} & \textbf{0.001} & \textbf{0.001} &
                         \textbf{0.031} & \textbf{0.036} & 0.142 & \textbf{0.031} & 0.211 & 0.165 \\
\textbf{Moirai\_Large}& \textbf{0.001} & \textbf{0.001} &
                         \textbf{0.017} & 0.069 & 0.135 & \textbf{0.028} & 0.169 & \textbf{0.045} \\
\addlinespace
\multicolumn{9}{c}{\textbf{Goods-Producing Industries}}\\
TimeGPT-1            & 0.317 & 0.534 & 0.490 & 0.831 & 0.268 & 0.463 & 0.443 & 0.978 \\
Chronos\_Small       & 0.174 & 0.390 & 0.917 & 0.258 & 0.188 & 0.587 & 0.596 & 0.519 \\
Chronos\_Base        & 0.184 & 0.396 & 0.855 & 0.116 & 0.209 & 0.635 & 0.678 & 0.584 \\
Chronos\_Large       & 0.164 & 0.901 & 0.560 & 0.431 & 0.178 & 0.875 & 0.836 & 0.622 \\
\textbf{Moirai\_Small}& 0.080 & \textbf{0.042} & \textbf{0.029} & \textbf{0.011} & 0.212 & 0.397 & 0.323 & 0.094 \\
\textbf{Moirai\_Base} & \textbf{0.025} & \textbf{0.001} & \textbf{0.024} & \textbf{0.008} & 0.072 & \textbf{0.032} & 0.125 & \textbf{0.024} \\
\textbf{Moirai\_Large}& \textbf{0.021} & \textbf{0.002} & \textbf{0.024} & \textbf{0.008} & 0.069 & \textbf{0.049} & 0.118 & \textbf{0.036} \\
\addlinespace
\multicolumn{9}{c}{\textbf{Services Industries}}\\
TimeGPT-1            & 0.443 & 0.614 & 0.237 & 0.100 & 0.273 & 0.615 & 0.194 & 0.701 \\
Chronos\_Small       & 0.257 & 0.136 & 0.233 & 0.790 & 0.279 & 0.167 & 0.362 & 0.382 \\
Chronos\_Base        & 0.183 & 0.126 & 0.278 & 0.496 & 0.197 & 0.188 & 0.268 & 0.368 \\
Chronos\_Large       & 0.287 & 0.137 & 0.971 & 0.369 & 0.315 & 0.150 & 0.279 & 0.371 \\
\textbf{Moirai\_Small}& \textbf{0.039} & \textbf{0.015} & 0.233 & 0.695 & 0.069 & \textbf{0.047} & 0.090 & 0.207 \\
\textbf{Moirai\_Base} & 0.054 & \textbf{0.039} & \textbf{0.005} & \textbf{0.020} & 0.097 & 0.121 & 0.121 & 0.214 \\
\textbf{Moirai\_Large}& \textbf{0.040} & \textbf{0.025} & \textbf{0.013} & \textbf{0.029} & 0.068 & 0.074 & 0.076 & 0.231 \\
\bottomrule
\end{tabular}
\caption{Diebold–Mariano two-sided \emph{p}-values comparing each TSFM with Persistence ($p_P$) and ARIMA ($p_A$) across four periods. Bold numbers denote $p<0.05$.}
\label{tab:forecast_Diebold}
\end{table}

The tables show the performance of the Moirai models, especially Moirai-1.1-R-base (Moirai Base) and Moirai-1.1-R-large (Moirai Large), indicating strong performance. These models halve typical errors relative to traditional baselines. Conversely, orange and red areas concentrate around Persistence and ARIMA in several sectors. TimeGPT-1 and Chronos-t5 models consistently occupy an intermediate position. The accompanying table presents the mean rank of RMSE for each model across four sectors, allowing for simple identification of the best models in this experiment.

\subsubsection{Forecast Analysis During Stable Phases}
During the three quiet years leading up to the pandemic (2017Q1 – 2019Q4), economic patterns were remarkably predictable. This predictability resulted in low forecast errors across all sectors, with evaluation metrics consistently indicating strong accuracy. Traditional forecasting algorithms thrived under these orderly conditions. The Persistence model, which simply projects the most recent value, proved surprisingly effective when conditions remained stable. ARIMA models, particularly adept at detecting seasonal patterns, performed exceptionally well in sectors, aligning with farming and export cycles.

However, the Chronos and Moirai models surpassed these traditional counterparts in four sectors. Their advantage stemmed from their ability to identify and leverage additional predictable data for incremental improvements. These models effectively captured underlying patterns, potentially related to cyclical behaviour, without overfitting to the period's general stability.

By contrast, the globally trained TimeGPT-1, whose strength normally lies in drawing on vast cross-domain context, had little additional information to exploit in such a calm and peaceful environment and therefore delivered respectable but middling accuracy. In short, the era’s predictability rewarded straightforward, transparent methods, leaving more sophisticated architectures only marginal room to demonstrate their advantages.

\subsubsection{Forecast Analysis of During Shocks}
The COVID-19 shock (2020Q1 – 2022Q4) ruptured the steady rhythms on which most forecasters relied. Forecast errors spiked across the board, and even algorithms with a reputation for versatility struggled. Tables \ref{tab:forecast_stacked_vertical} and \ref{tab:forecast_Diebold} show that the pandemic's impact still leaves a measurable track during the COVID-19 period in these two sectors. Some models show error peaks during the COVID-19 pandemic.

These spikes, doubling in some cases, highlight how the extreme, sudden shifts in economic activity during the pandemic strained even the most sophisticated forecasting approaches. This pushed Persistence to the bottom of the rankings. ARIMA also performed less well on National GDP but had fewer errors in Primary Industries, where commodity-driven swings were less violent. Chronos and Moirai proved the most resilient, preserving green scores across sectors. They absorbed the shocks with a level of grace unmatched by simpler rivals, their attention mechanisms and multi-scale encoders proved capable of rapid model adaptation. In contrast, TimeGPT-1 struggled to digest the unprecedented data, often sliding into the red. The experience made clear the challenges in extreme situations.

\begin{figure}[htbp]
\centering
\includegraphics[width=0.9\textwidth]{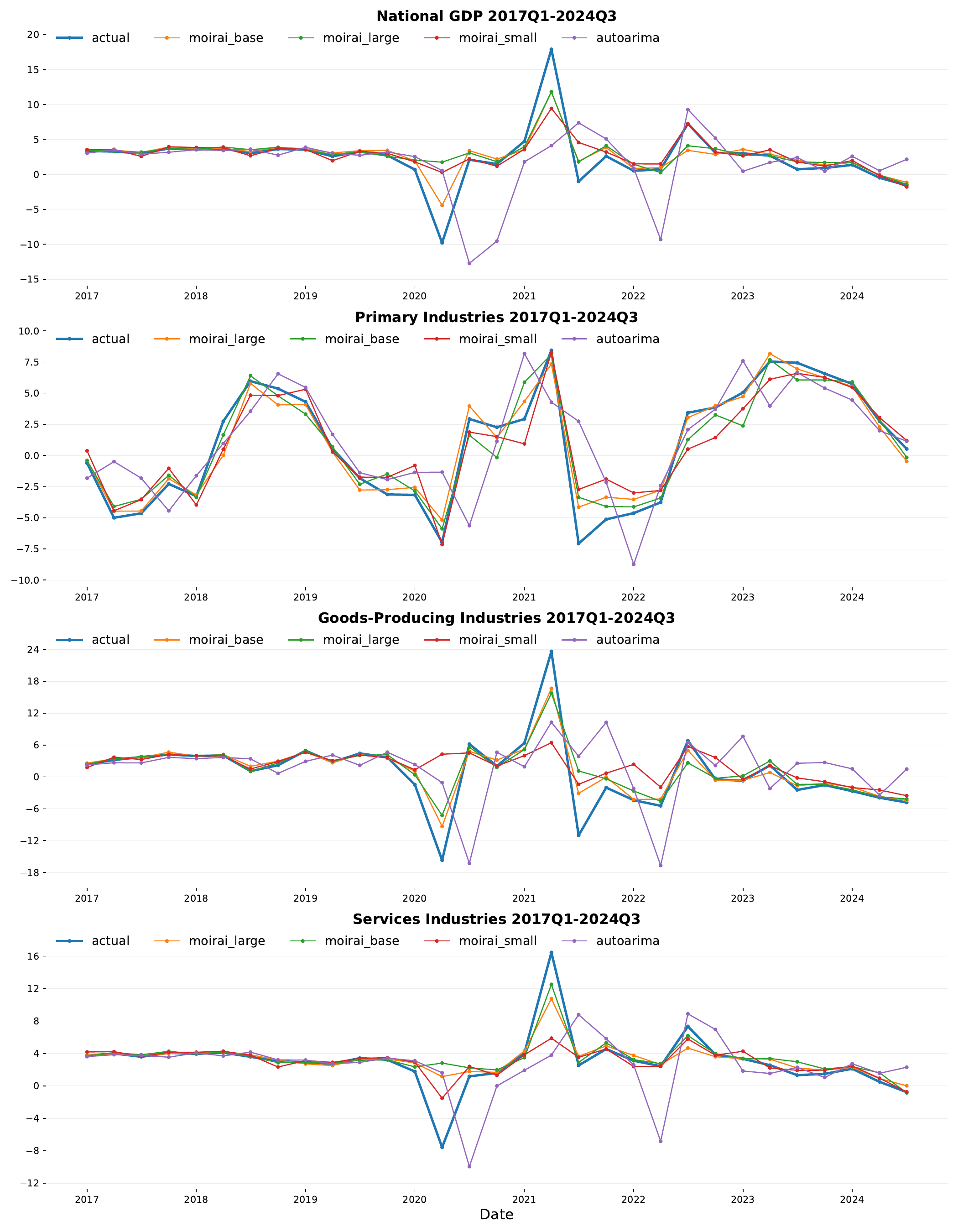}
\caption{Actual and forecasting values for the quarterly annual percentage change from 2017Q1 to 2024Q4.}
\label{fig:overall_percentage_chage}
\end{figure}

The COVID-19 shock affected forecasting accuracy over different periods, as shown in Figure \ref{fig:overall_percentage_chage}. Showing absolute error units indicates the magnitude of errors, while the percent-change table immediately reveals which models and metrics saw the biggest proportional impacts in the Post-COVID-19 period. This experiment thus provides a clear, reproducible way to assess how sudden shifts occur.

\subsubsection{Forecast Analysis Post Instability}
Post-COVID-19 (2023Q1 – 2024Q3) analysis pinpoints how forecasting accuracy shifted by clearly dividing the horizon into two phases, Pre-COVID and Post-COVID. Comparing the error metrics across these windows reveals the damage and relative resilience of the models. Differences between models are significantly impacted by the dramatically widened error margins during the crisis. The pandemic's aftershocks echo in the data-generating process, as Post-COVID error metrics are distinctly higher than Pre-COVID. However, as the immediate Post-COVID period recedes, forecasting accuracy has rebounded. Most metrics have fallen by two-thirds from their COVID peaks, and ``greens'' (indicating good performance) re-emerge in every column. Persistence models, often brittle in the face of extreme shocks, now produce errors that are no longer extraordinary, becoming respectable again.

Despite this general recovery, Moirai models consistently deliver leading accuracy across the four sectors. They reclaim leadership and regain dominance by capturing residual asymmetries that persist after the shock, thereby sharpening distinctions among the other models. ARIMA is rarely at the top of the leaderboard but stays within a comfortable margin of the front-runners. TimeGPT-1 continues to trail the specialist models, lacking fine-tuned sensitivity in these domain-specific economic series. Thus, the zero-shot forecasts of Moirai and Chronos models are holding the line during these pandemic-induced spikes. These models demonstrate an ability to handle unprecedented incidents, ruptures, and behavioural swings.

\subsubsection{Summary}
Overall, classical and statistical baseline models such as persistence and ARIMA, exhibit increased errors across most metrics. In contrast, TSFMs demonstrate considerably greater resilience. Their MAE and MASE remain close to zero, and RMSE frequently shows improvement, suggesting that extensive pre-training enables them to adapt to new levels. TimeGPT‑1 is notable for significant error increases, particularly concerning National GDP. The Moirai models represent a middle ground; they are better at capturing scale shifts with lower RMSE but still experience increases in MAE and SMAPE. This analysis was conducted over 26 years across four sectors, evaluating performance using multiple error metrics (MAE, RMSE, SMAPE, MASE).

\begin{figure}[htbp]
\centering
\includegraphics[width=0.9\textwidth]{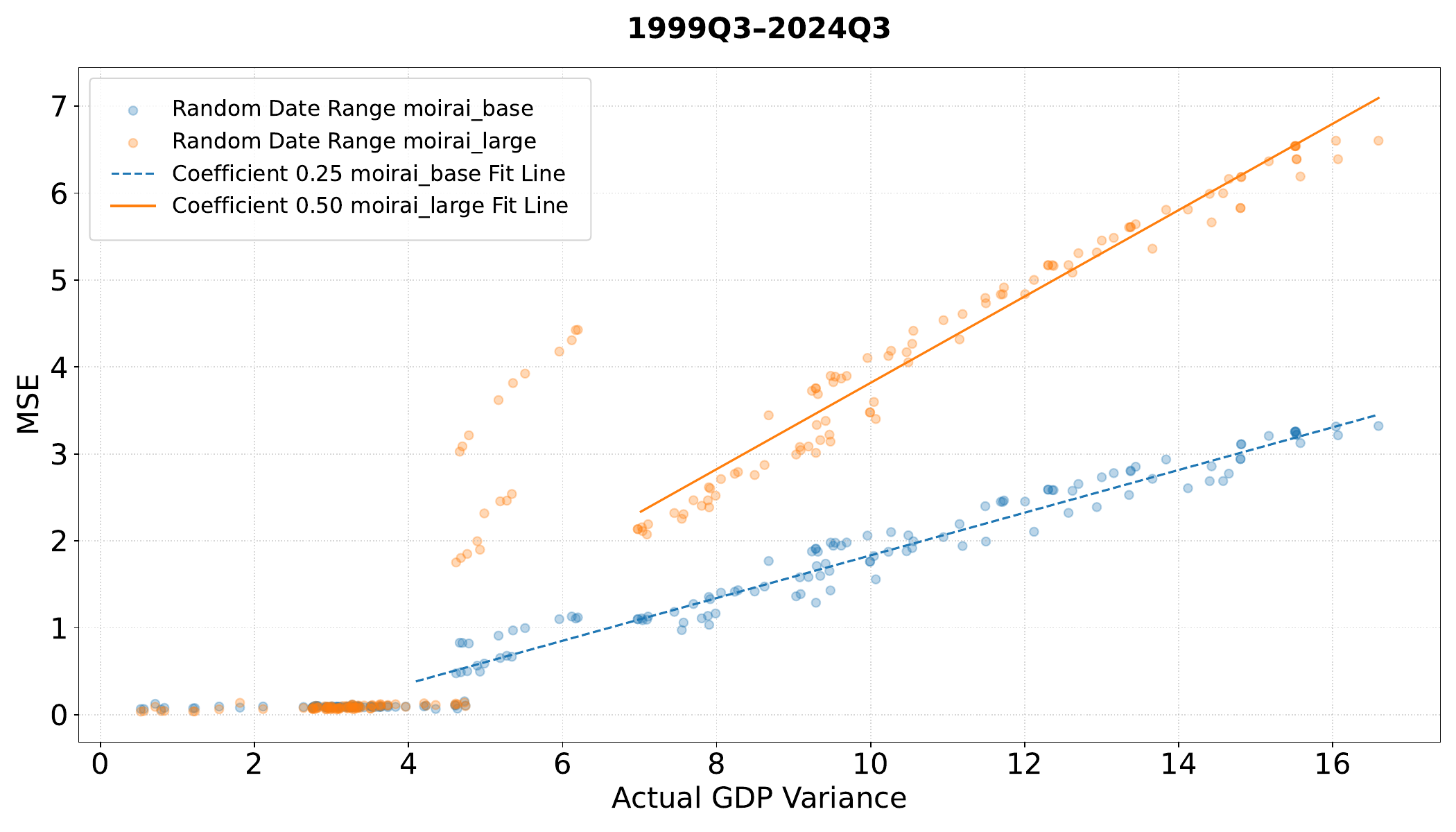}
\caption{A regression plot of actual GDP variance against MSE over 26 years.}
\label{fig:GDP_RMSE_variance_comparison_single}
\end{figure}

Moirai-1.1-R-base (Moirai Base) and Moirai-1.1-R-large (Moirai Large) demonstrate strong consistency with low mean ranks of RMSE across four sectors over time. The COVID-19 period, reflecting a difficult time for models adapting to sudden structural shifts, presented a significant challenge. As markets begin to normalise Post-COVID, the differences in handling volatility become clearer. 

The regression analysis Figure \ref{fig:GDP_RMSE_variance_comparison_single} highlights this: Moirai Base appears to cope with macroeconomic turbulence, such as the COVID-19 shock, more robustly than Moirai Large. To quantify this, we drew 200 random sub-periods from the 1999Q3–2024Q4 sample (each window containing at least 10 quarters), computed the variance of realised GDP growth for every window, and paired it with the model’s out-of-sample RMSE. Plotting variance (x-axis) against RMSE (y-axis), therefore yields 200 (variance, RMSE) points for each model. A linear regression fitted to these clouds reveals a markedly flatter slope for Moirai Base than for Moirai Large, indicating that the large model’s error climbs much faster as volatility increases. In practical terms, Moirai Base maintains accuracy when conditions become volatile, whereas Moirai Large deteriorates more sharply, making the base model the more advantageous choice during periods of economic instability.

\subsection{Forecast Benchmarking Against State-of-the-art}
We compare our best model, Moirai Base, with the RBNZ's LSboost and Factor models, using the quarterly percentage change for overall National GDP forecasts.

\begin{figure}[htbp]
\centering
\includegraphics[width=0.9\textwidth]{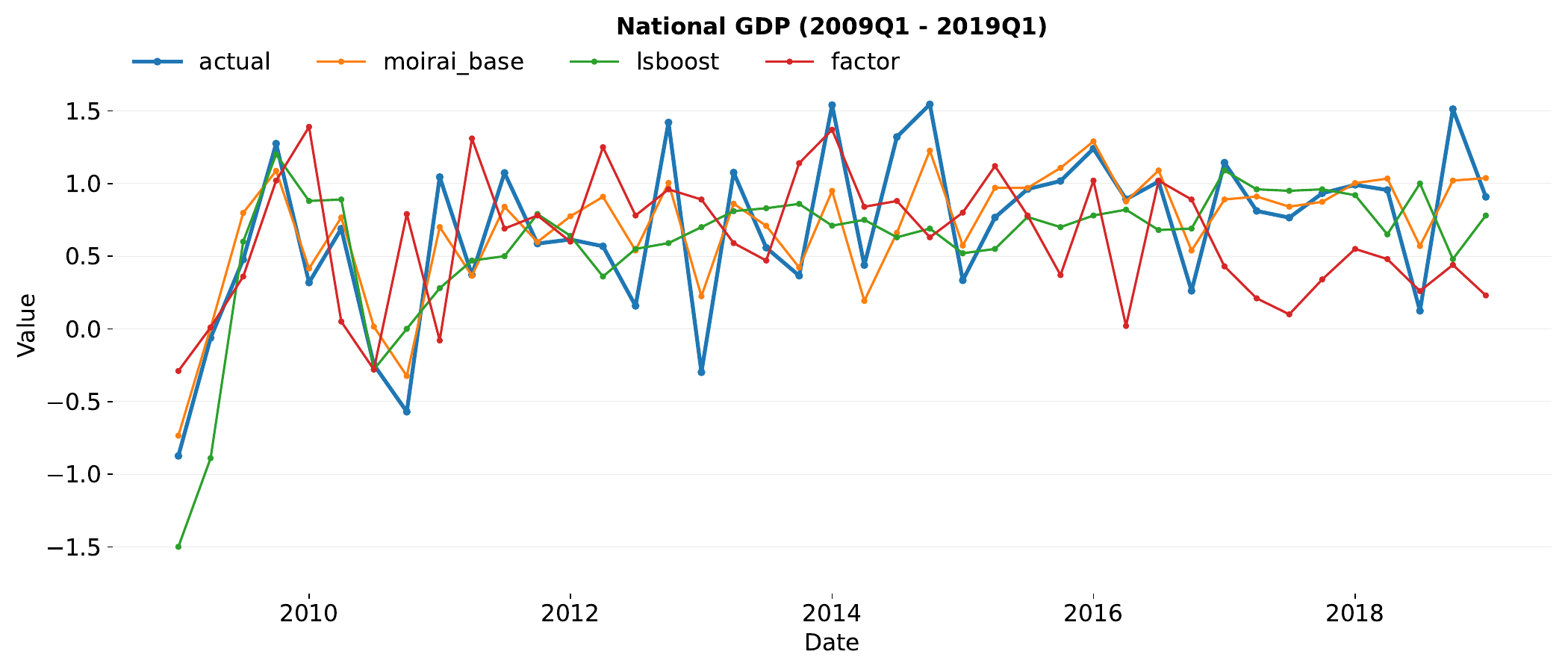}
\caption{Comparisons of the Moirai Base Model with LSboost and Factor Model using Reserve Bank Nowcasts.}
\label{fig:Comparison_with_RZ}
\end{figure}

We extracted forecasts from the RBNZ's LSboost (Gradient boosting) and Factor model, observing that they are quarterly percentage changes. Consequently, we converted the national GDP data from a quarterly annual percentage change to a quarterly percentage change to match the forecast format. For comparison in Figure \ref{fig:Comparison_with_RZ}, we also used Moirai Base to forecast quarterly percentage change values. The table includes Diebold Mariano two-sided tests computed against an ARIMA model for the period 2009Q1 - 2019Q1, evaluating the RBNZ's published LSboost and Factor GDP nowcasts \cite{Richardson2021Nowcast}. Table \ref{tab:dm_pvalues} presents forecast accuracy comparisons based on RMSE and Diebold–Mariano (DM) tests. Moirai Base yields the lowest RMSE, indicating the best point-forecast performance overall. The DM statistics show that Moirai’s errors are significantly smaller than those of a benchmark ARIMA model. Still, the differences from the RBNZ’s LSBoost and Factor models are not statistically significant at conventional levels. Thus, while Moirai demonstrably outperforms ARIMA, it performs on par with LSBoost and Factor, an important finding that highlights their comparable predictive accuracy.

\begin{table}[ht]
\centering
\begin{tabular}{@{}l rr@{}}
    \toprule
    \multicolumn{3}{c}{\textbf{Comparison}} \\
    \cmidrule(lr){1-3}
    \textbf{Model comparison} & \textbf{RMSE} & \textbf{$p$-value} \\
    \midrule
    Moirai Base              & 0.2692 & 0.0014 \\
    LSBoost \cite{Richardson2021Nowcast} & 0.4876 & 0.2389 \\
    Factor \cite{Richardson2021Nowcast}  & 0.6328 & 0.4936 \\
\bottomrule
\end{tabular}
\caption{Diebold-Mariano test $p$-values comparing with models of RBNZ}
\label{tab:dm_pvalues}
\end{table}

\section{Discussion}
\subsection{TSFM Zero-Shot Effectiveness in Macroeconomic Forecasting}
Based on \textbf{RQ1}, empirical studies reveal a clear hierarchy among state-of-the-art TSFMs in a zero-shot setting. These TSFMs represent promising steps toward universal forecasting. Empirically, Moirai currently demonstrates the most consistent accuracy on macro and industry series, exhibiting the lowest scaled errors and highest ranks. This is likely attributable to task-specific design and massive pre-training \cite{Moirai2024}. Chronos also performs strongly, especially on monthly and other high-frequency series matching state-of-the-art zero-shot accuracy with only minor configuration tweaks \cite{Chronos2024}. By contrast, although TimeGPT offers broad cross-domain versatility, it trails leading alternatives on several specialised economic benchmarks \cite{TimeGPT2024}. Both models, however, can significantly narrow this gap when they are fine-tuned or adapted in context with domain-specific economic and industry datasets \cite{das2024incontextfinetuningtimeseriesfoundation}. Overall, these models generalise better across domains than traditional local models. However, it is important to be aware of their limitations and select model size and type accordingly. Because zero-shot forecasts rely solely on patterns learned during pre-training, they cannot adapt when the data-generating process shifts to a new regime not represented in the training corpus.

Regarding the research question, we can address that TSFMs are indeed a powerful new forecasting tool whose zero-shot forecasts can rival and even surpass classical and deep-learning baselines. Nevertheless, a limitation arises when evaluating a zero-shot forecast on the latest data vintage while a baseline was trained on older or different data. In such cases, the comparison involves forecasting different targets, making meticulous vintage control and strict alignment of transformations crucial to avoid a mismatch.

\subsection{Effectiveness of TSFMs in Zero-Shot Forecasting during Shocks}
For \textbf{RQ2}, we observe that:
TSFMs are not immune to the effects of abrupt market changes, they allocate additional representational capacity to capture sudden spikes and crash patterns once these appear in the training data stream. In experiments, this manifests as the zero-shot forecast overshooting immediately after a shock subsides, followed by a glide path before error metrics converge to their pre-crisis baselines. Pre-trained models exhibit an even stronger lag because their universal embeddings adapt more slowly to new domain-specific data, such as the shifts seen during the pandemic. Consequently, Post-COVID forecasts inherit a systematic bias, and their prediction intervals remain excessively wide. This phenomenon has also been documented in earlier crisis episodes, where models struggled to distinguish between statistical noise and lasting shifts with higher penetration. Nevertheless, zero-shot forecasts are not flawless. The initial quarters following an unprecedented shock often exhibit residual biases in predictions, especially when the disruption’s characteristics diverge significantly from the model’s learned priors or involve novel policy responses outside the training distribution. In such cases, light Post-event calibration can be beneficial. It is important to recognize that zero-shot does not equal shock-proof. Because no local fine-tuning occurs, TSFMs can carry a small residual bias in the few Post-shock quarters, potentially overreacting if the current crisis deviates significantly from patterns seen in pre-training.

Addressing the research question, the findings suggest that one cannot rely on the zero-shot forecast alone. While TSFMs offer a theoretical advantage in leveraging cross-domain patterns and adaptive attention, modest fine-tuning or bias correction using new data can markedly improve forecast accuracy over the long run without eroding the model's broad generalization capabilities. This highlights the importance of modest adaptations to effectively handle novel extreme events.

\subsection{Zero-Shot TSFMs vs Domain-Specific Models for Macroeconomic Forecasting}
\textbf{RQ3} sought to investigate whether fully zero-shot TSFMs can compete with, or even replace, the bespoke multivariate systems employed by central banks. Diebold-Mariano tests on real-time New Zealand GDP forecasts demonstrate that Moirai-1.1-R-base is statistically superior to an ARIMA benchmark. However, it is only comparable to the RBNZ LSBoost ensemble and dynamic-factor model, not significantly better. Consequently, while the RBNZ's suite retains a narrow statistical edge, the performance gap has effectively closed.

The decisive factor in this comparison is information breadth. LSBoost and the factor model incorporate hundreds of carefully curated covariates, including business-confidence surveys, commodity prices, export receipts, and high-frequency trackers. These capture cross-sectional signals that a univariate zero-shot model inherently lacks. Despite this handicap, Moirai’s relative success stems from its architecture and pre-training strategy: a hierarchical transformer trained on millions of heterogeneous series. This training enables it to learn universal priors for seasonality, sudden shifts, and global disturbances. These priors allowed Moirai to maintain low errors through the COVID-19 shock periods, during which many traditional models required re-estimation.

Addressing the research question, the demonstration that TSFMs can match the RBNZ’s sophisticated multivariate machinery emphasise values for agencies with limited analytical resources. Furthermore, because Moirai’s performance remains stable during sudden shocks, embedding its forecasts in early-warning dashboards would bolster risk monitoring and help policymakers respond more swiftly to economic turning points. In economies where external shocks propagate quickly, such robustness is particularly valuable for setting prudent fiscal buffers, calibrating macro-prudential tools, and stress-testing contingency plans.

\subsection{Limitations and Future Works}
Although our back-tests suggest that TSFMs set a competitive baseline for zero-shot GDP forecasting, those performances are contingent on the stability of the underlying time series and economic environment. This study also acknowledges the possibility that time series from our experimental domain may have been included in the training datasets for the various TSFMs which may have influenced the accuracy results. A further limitation is that our study did not consider multivariate capabilities of TSFMs, which we leave to future work.
 
Furthermore,  we plan on expanding our future work by moving beyond zero-shot evaluation to explore a fine-tuning approach for re-training models with plausible exogenous shocks. This expansion will also involve extending evaluations to other economies and time periods for accuracy improvement, and applying a quarterly intercept correction to trim forecast errors without compromising cross-domain generality. Lastly, reporting full predictive distributions via density forecasts and coverage metrics for assessing these models deliver calibrated probabilities as well as sharp point forecasts.

\section{Conclusion}
This research delivers one of the first systematic zero-shot evaluations of TSFMs for macroeconomic indicators forecasting. Without any fine-tuning, several TSFM variants, including TimeGPT, Chronos, and Moirai, were tasked with predicting the quarterly year-on-year growth rates of four headline series: National GDP, Primary Industries, Goods-Producing Industries, and Services Industries. Each model operated in a purely univariate setting, eliminating bespoke econometric specifications and long local training histories. Point forecasts and predictive interval calibration were benchmarked against classical/statistical baselines and operationalised models from large institutions.

The Moirai variants outperformed Persistence and ARIMA across several horizons and matched industry benchmarks. Crucially, TSFMs remained broadly resilient during the COVID-19 structural break, yet the findings revealed small and systematic post-shock biases that suggest a need for multivariate extensions that incorporate other indicators. These results show that carefully engineered TSFMs can internalise rich economic dynamics and deliver actionable forecasts for macroeconomic monitoring and strategic planning. At the same time, clear directions exist for future work targeting, multivariate integrations, fine-tuning, ensemble blending, and rigorous stress testing across additional crisis scenarios.

\bibliographystyle{unsrtnat}
\bibliography{main}  

\appendix

\clearpage

\section{Appendix A}
\begin{figure}[htbp]
  \centering
  \includegraphics[width=0.75\textwidth]{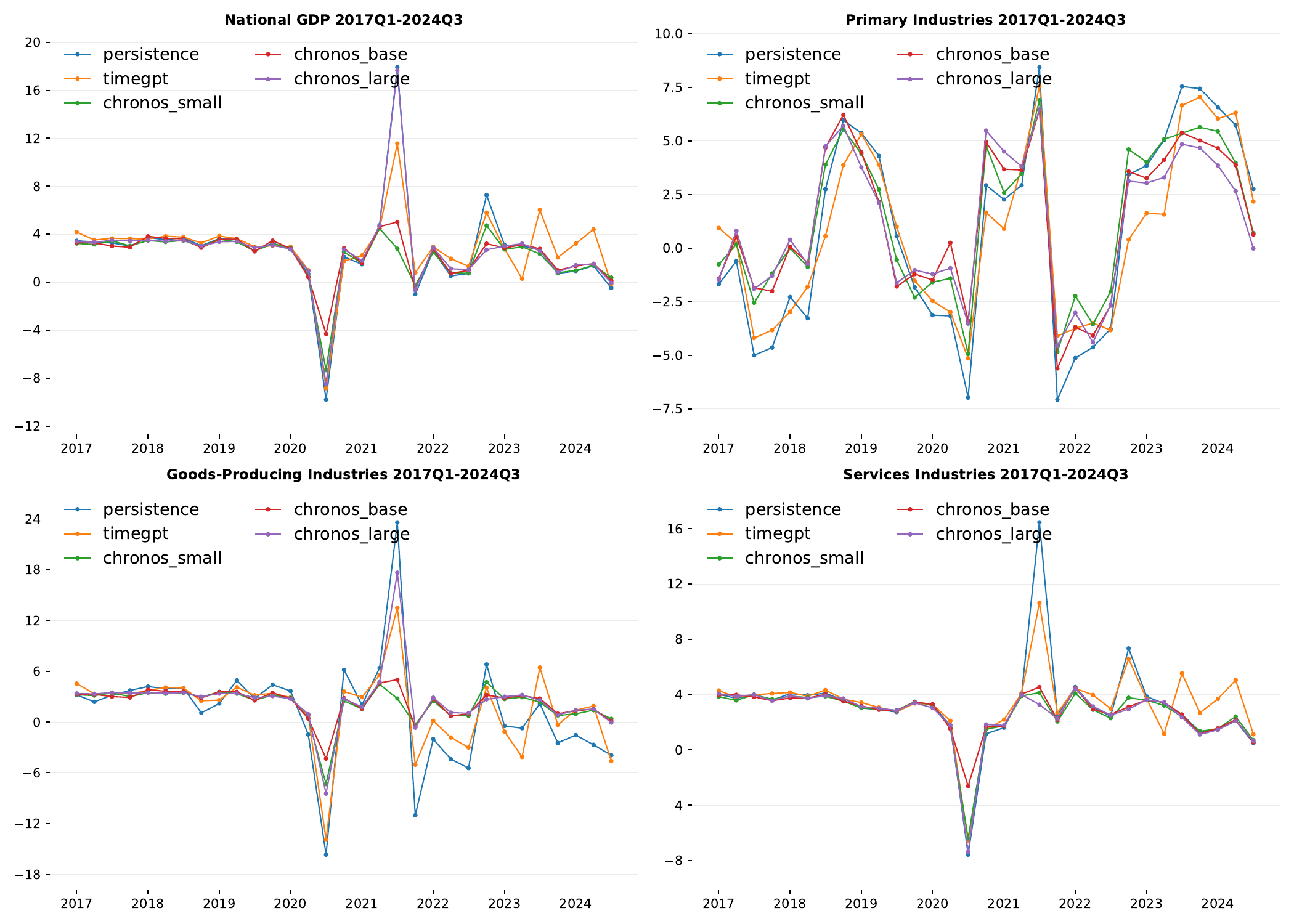} 
  \caption{Forecasting accuracy results: quarterly annual percentage change by remaining models (2017Q1 - 2024Q3)}
  \label{fig:other_models_results}
\end{figure}

Figure \ref{fig:other_models_results} shows the quarterly annual percentage change in the forecasting accuracy of other models for National GDP, Primary Industries, Goods-Producing Industries, and Services Industries over the period 2017Q1 - 2024Q3.

\end{document}